\documentclass[runningheads,a4paper]{llncs}
\usepackage{url}
\urldef{\mailsa}\path|{alfred.hofmann, brigitte.apfel, ursula.barth, christine.guenther,|
\urldef{\mailsb}\path|ingrid.haas, frank.holzwarth, anna.kramer, leonie.kunz, nicole.sator,|
\urldef{\mailsc}\path|erika.siebert-cole, peter.strasser, lncs}@springer.com|

\usepackage{hyperref}
\hypersetup{
    colorlinks,
    linkcolor={red!50!black},
    citecolor={blue!50!black},
    urlcolor={blue!80!black}
}

\usepackage{times}
\usepackage{wrapfig}
\usepackage{amsmath}
\usepackage{amssymb}

\usepackage{wrapfig}

\usepackage{graphicx} % support the \includegraphics command and
\usepackage{epsfig}
\usepackage{epstopdf}
\usepackage[monochrome]{color}

\usepackage{multirow}
\usepackage{mathrsfs}
\usepackage{algorithm}

\usepackage[noend]{algpseudocode}

\makeatletter

\newcommand*{\rom}[1]{\expandafter\@slowromancap\romannumeral #1@}

\makeatother

\begin{document}
\mainmatter              % start of the contributions

\title{Ensemble of Multi-sized FCNs to Improve White Matter Lesion
  Segmentation}

\titlerunning{Ensemble of Multi-sized FCNs}

\author{
  Zhewei Wang\inst{1}%
\and Charles D. Smith\inst{2}
\and Jundong Liu\inst{1}
}

\authorrunning{*, *, *, *}
\institute{
  % School of Electrical Engineering and Computer Science, Ohio University, Athens OH
  School of Electrical Engineering and Computer Science, Ohio
  University \and Department of Neurology, University of Kentucky }

\toctitle{Ensemble of Multi-sized FCNs}
\tocauthor{Wang, Smith, Liu}

\maketitle

\begin{abstract}
  In this paper, we develop a two-stage neural network solution for
  the challenging task of white-matter lesion segmentation. To cope
  with the vast variability in lesion sizes, we sample brain MR scans
  with patches at three different dimensions and feed them into
  separate fully convolutional neural networks (FCNs). In the second
  stage, we process large and small lesion separately, and use
  ensemble-nets to combine the segmentation results generated from the
  FCNs.  A novel activation function is adopted in the ensemble-nets
  to improve the segmentation accuracy measured by Dice Similarity
  Coefficient. Experiments on MICCAI 2017 White Matter
  Hyperintensities (WMH) Segmentation Challenge data demonstrate that
  our two-stage-multi-sized FCN approach, as well as the new
  activation function, are effective in capturing white-matter lesions
  in MR images.

\end{abstract}

\section{Introduction}
 
% \verb+{MS lesion}+
%Multiple Sclerosis (MS) is a central nervous system disease in which
%the protective sheath (myelin) of nerve fibers are mistakenly attached
%by the immune system. Such attacks cause focal inflammation and result
%in scar tissues (lesions), which can be observed through Magnetic
%Resonance Imaging (MRI).
% \verb+{Segmentation is important}+
Multiple Sclerosis (MS) may result in lesions within patients' white
matter tissues, which can be observed through Magnetic Resonance
Imaging (MRI).  Identifying and measuring the spatial and temporal
disseminations of lesions are key components of diagnostic criteria
for MS.
% \verb+{manual is tedious, so automatic}
% As manual segmentation of MS lesions is time-consuming and prone to
% large intra- and interexpert variability, significant research efforts
% have been seen in the past 15 years or so in developing automatic
% segmentation solutions.
%
% \verb+{Traditional methods include supervised and unsupervided}
Traditional approaches to segment MS lesions include the utilization
of supervised classification
\cite{anbeek2004automatic} %,younis2007ms,harmouche2006bayesian,geremia2011spatial}
or unsupervised clustering
\cite{Warfield2015323,souplet2008automatic,liu_2009_mmbia} %
                                           % ,garcia2011trimmed,van2001automated}
to separate lesions from the normal brain tissues, where the former
are commonly modeled as either an additional class or the outliers to
the latter.
% When no training samples are available, deformable models have also
% been employed to delineate lesion boundaries \cite{ZHAO201894}.  ,
% with impressive results reported.

%   \verb+{Deep learning emerges as new and powerful paradigm}
In recent years, deep learning models, especially convolutional neural
networks (CNN), have emerged as a new and more powerful paradigm in
handling various artificial intelligence tasks, including image
segmentation.
% \verb+{Start with patch-based, taking patch label as voxel lable} \verb+{FCN
%   is powerful} Prior to 2015, patch-wise CNN architectures were
% popular, where patches of the training samples are extracted to
% train a deep classification CNN. In the testing stage, each pixel
% takes the label of its surrounding patch, estimated through the
% network.  In this way, the image segmentation task becomes a dense
% patch-wise classification for each extracted image patch.
% {\color{black} While impressive results were produced, patch-wise
%   CNNs commonly ignore the relationship between adjacent pixels, and
%   an appropriate patch size is often difficult to be set
%   automatically.  In addition, the gap between global information
%   (patch) and local membership (pixel) imposes an inherent
%   limitation to this group of solutions. } To add context into the
% patch/pixel labeling procedure, several recent works
% \cite{farabet2013learning,kamnitsas2017efficient,moeskops2016automatic,song2015accurate,yang2017cascade}
% explore the so-called multi-stream paradigm, feeding networks with
% patches of different scales or different angles. Such integrations,
% However, cannot eliminate the inherent drawbacks of patch-wise CNNs.
%
%Taking a semantic approach,
Being able to process information from various spatial scales, the
fully convolutional networks (FCN) \cite{long2015fully} and its
variants \cite{ronneberger2015u,vnet_2016}
% including U-Net \cite{ronneberger2015u} and V-Net \cite{vnet_2016},
have gained great % increasing
popularity in recent years.
% For medical images, the U-Net and 3D U-Net are arguably the most
% well-known FCN variants.
U-Net \cite{ronneberger2015u} is arguably the most well-known FCN
model in medical image analysis. It utilizes a deep CNN to encode
discriminative features of the training images and relies on a
deconvolution decoder to integrate the features together in producing
the segmentation results.
% \verb+{for lesion type segmentation, because of the variabilit of lesion size
%   multi-stream approaches have been utilized. }

While U-net and its extensions
\cite{vnet_2016,3d_unet,chen_isbi_2017,chen_mlmi_2017} are proven
effective to segment fixed sized objects (e.g., organs or cells), they
may not fare well for MS lesions, especially when the network is
designed to optimize {\it Dice Similarity Coefficient} (DSC).
% (or Intersection over Union (IoU) in computer vision literature).
% and the basis for objective function.
% This is in part due to the vast variability existing in lesion sizes
% -- large lesions can easily contain thousands of voxels, while the
% tiny ones may be as small as a single voxel.
MS lesions have a huge variability in size -- large lesions can easily
contain thousands of voxels, while many tiny ones are as small as only
1-2 voxels. As DSC is computed based on all foreground voxels, large
lesions are more important and tend to be treated with favor, while
small lesions could be overlooked without much
penalty.  %Overlooking the small lesions does not hurt much in
% Dice.
% and the latter may be easily overlooked.
Moreover, as DSC is not differentiable and therefore cannot be
directly used for gradient descent, many neural network models
\cite{vnet_2016,sudre2017generalised,ekanayake2018generalised,drozdzal2016importance}
take a probabilistic version of DSC, as an approximation to the
discrete DSC, in their objective functions.
% , hoping to obtain high DSC value in the testing stage.
Such approximation, however, deserves careful scrutiny, as the
theoretical gaps between discrete optimizations and continuous
optimization are generally difficult to overcome.

% Dice, as a discrete measure, cannot be directly used as
%an objective function in gradient descent-based NNs. A number of work
%use its probabilistic version (C-Dice) as an approximiation, but
%theoretical gaps exist between the optimizations of C-Dice and the
%discrete Dice (D-Dice) -- optimizing C-Dice doesn't necessarily lead
%to an improvement in D-Dice.

% Dice is a commonly used evaluation metric and its probabilistic
% version has been utilized as the objective functions in some network
% solutions. Two limitations may emerge when such network setting is
% applied to capture lesions: 1) because each voxel has equal
% importance in Dice, Dice favors to capture large lesions more
% accurately, and small lesions (because they contain small number of
% voxels) tend to be overlooked; 2) there are theoretical gaps between
% probabilistic Dice (C-Dice) and binary Dice (D-Dice) -- optimizing
% C-Dice doesn't guarantee an improvement to D-Dice.

% An FCN designed with a large input size would tend to overlook small
% sized lesions, while a small-patched FCN is prone to generate false
% positive labeling.  {\color{red} Regarding a remedy, we believe 1)
%   large and small lesions should be processed separately; 2) FCNs
%   targetting at different input sizes can be used to best capture the
%   respective sized lesions.} Where to put MICCAI 2017 Challenge
% papers? {\color{red} When it comes to combine the segmentation results
%  from multi-stream FCNs..}
%% Designing FCNs with different input sizes and using them to target
%% large and small lesions separately, would provide a remedy.

In this paper, we look into these two issues and propose a remedy
based on a two-stage multi-sized FCNs architecture.
% and use different sized FCNs to better capture them.
To cope with the lesion variability issue, we process large and small
lesions separately, and use ensemble nets to combine the segmentation
results from different sized FCNs. The contributions of individual
FCNs to the overall segmentation are automatically determined.
% sample brain MR scans
% with patches at three different sizes, feeding them into separate
% U-Net-like NNs. The segmentation results are then combined in the
% second-stage via an ensemble net, in which,
% The contributing weights of the individual segmentation from the
% first-stage FCNs are automatically determined.
To bridge the gap between discrete and continuous DSCs, we tackle the
issue from the activation function perspective,
% explore this issue through the activation function perspective,
and propose a new activation function to facilitate the network
training and improve the segmentation accuracy.
%
% as the replacement of the sigmoid function.
%
% we propose a two-stage multi-sized FCNs architecture to enhance the
% segmentation of MS lesions. To cope with the variability in lesion
% sizes, we sample brain MRI scans with patches at three different
% sizes, feeding them into separate U-Net-like NNs. The segmentation
% outputs produced in the NNs are consequently fused in the second
% stage, through an ensemble-net, to construct a combined labeling
% result. Contributions from the individual segmentations are
% automatically decided based on a novel activation function.

%Experiments on MICCAI 2017 Challenge data demonstrate that our
%two-stage multi-size settings, as well as the ensemble net, are
%effective in producing accurate and robust segmentation results for
%% MRI white-matter
%MS lesions in MRI.

% \input method_short

% \section{Method}
\section{Dice as evaluation metric and objective function: Issues}
\label{section2}
% {\bf Dice as evaluation metric and objective function}
% {\bf Discrete Dice and contenous Dice} 
% Definition of B-Dice
Let $S$ be the segmentation result produced by a solution and $R$ be
the ground truth, both of which are binary maps defined on the entire
image domain. DSC relies on the similarity of $S$ and $R$ to measure the
segmentation accuracy:
\begin{equation} \label{eq:DSC}
  DSC=\frac{
    2|{S}\cap R|}{|S|+|R|}
\end{equation}

{\bf Segmentation Biases} As equal weights are assigned to all
foreground voxels in $S$ and $R$,
% While DSC is one of the major goals when design segmentation
% solutions, Because each voxel counts,
seeking higher DSC would strive to capture large lesions, but tend to
overlook small lesions, resulting in false-negative (type II) errors.
When the input size of an FCN is set to the entire image or a large
sub-volume, such effect can be easily observed. This situation can be
regarded as a special type of data imbalance.
%Pooling operation along the deep layers also plays a role. 
When the inputs are set to small sub-volumes, global information would
be limited within each input, and false positive (type I) error are
prone to be generated. In addition, for the small patches that are
mostly immersed within a large lesion, they may contain more lesion
voxels than non-lesional ones. This leads to a different type of input
imbalance, which could also result in erroneous segmentations. When
DSC is used as a segmentation evaluation metric, all these biases are
inherent to the system. To reduce them, processing the images with
sub-volumes of different sizes
% different sized subvolumes % sized FCNs
can potentially provide a remedy.
% This is the design rational behind our two-stage multi-sized network
% architecture, to be presented in next section.

%B-Cide can not be directly used as the objective function for NNs,
%because backpropogation requires function gradient, but
{\bf Discrete vs. Continuous Dice} Defined on binary maps, DSC in
Eqn.~\ref{eq:DSC} is not differentiable, therefore cannot be directly
used as the objective function for FCNs.
% B-Dice is not differentiable, so
In practice, a probabilistic or continuous version DSC, called Dice loss
\cite{vnet_2016}, has been used as an approximation to lead the
updates of segmentation. In Dice loss, segmentation $S$ is relaxed to
a probability map of real numbers between $0$ and $1$, and the loss is
computed as: 
\begin{equation} \label{eq:1}
\textrm{Dice loss} = -\frac{2\sum_i{s_ir_i}}{\sum_is_i+\sum_ir_i}
\end{equation}
where $s_i\in[0,1]$ is the label prediction at voxel $i$, and
$r_i\in \{0,1\}$ is the corresponding binary ground truth.
% Due to the relaxation, the optimization of Dice loss cannot refelect
% the value change of DSC well.  Improving C-Dice doesn’t necessarily
% improve B-Dice.

There exists, however, a theoretical gap between the discrete
optimization of DSC and the continuous optimization of Dice loss. An
illustration example is given as follows.
% Many ways to improve the approximation, one of which would be a
% better activation function （better approximation, numerically).
%
\begin{figure}[!htb]
\begin{center}
  \includegraphics[width=4in]{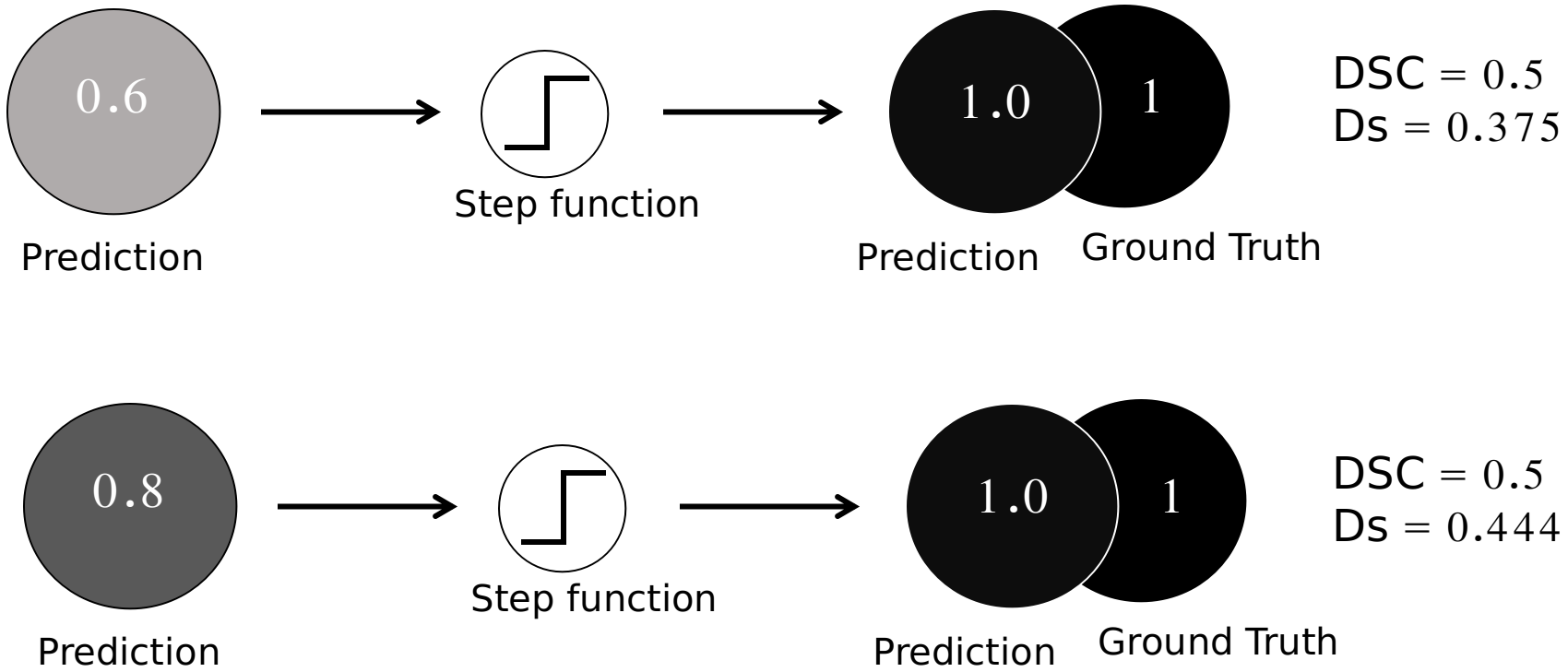}
\end{center}
\caption{An example to illustrate the gap between DSC and Dice loss
  (Ds). Refer to text for more details.}
% \caption{Illustration of a case that consists of two overlapped
% disks. A step function is used to map the prediction as binary
% values. {\color{red}{Illustration of a case that consists of two
% overlapped disks. In this case, the increase of intensity (from 0.6
% to 0.8) can boost Dice loss (from 0.375 to 0.444), but desn't
% improve DSC.}}}
\label{fig:dice_illustration}
\end{figure}
%
% For example,
Consider two overlapping disks, one is the ground truth segmentation
with intensity value of $1$, and the other is the prediction with
probabilities in the range of $[0, 1]$. Assume the overlapping area
between the ground truth and the prediction is half of the disk. To
calculate DSC, the prediction map needs to be converted to binary
through a Step function that uses $0.5$ as the threshold.
%defined as in equation \ref{eq:step}.
In Fig.~\ref{fig:dice_illustration}, two prediction maps are shown,
with the across-the-board probabilities of $0.6$ and $0.8$, respectively.
% (a) or Fig.~\ref{fig:dice_illustration} (b).
%  \begin{equation} \label{eq:step}
%    \chi(x)=\begin{cases}
%      1 \quad \phantom{\infty}\text{if}\,\, x \ge 0.5   \\
%      0 \quad \phantom{\infty}\text{if}\,\, x < 0.5
%    \end{cases}
%  \end{equation}
The upper case (prediction = $0.6$) has a Dice loss of $0.375$, and
that for the lower case (prediction = $0.8$) is $0.444$. The discrete
DSCs for them, however, are both $0.5$.

This example illustrates that decreases in the Dice loss do not
necessarily lead to decreases (or even changes at all) in the values
of DSC. This gap can pose serious difficulty in the network training
procedure. Our approach to bridge the gap is through an activation
function point of view, and the design will be presented in next
section.
% This observation leads to the design a new activation function in
% our two-stage network, to be presented in next section. 
% A related observation (not from this figure) is that
% activation functions that are closer to the step function would be
% able to provide a better approximation to the DSC. In other words,
% steep activation functions are generally preferreed in this regard.
% On the other hand, such steep function, if to be used as an activation
% function, should have a resonable capture range to facilitate
% convergence.
  % 
% from the convergence considertation, functions with very narrow
% capture range should also be avoided.
% {\bf With these two considerataions, I propose a new activitation
%   function for our ensemble net.}
 % 
  %
  % The threshold $0.5$ is key, and to better approximate the step function,
  % the activitation function should be as close as step 
 % 
%
  % To alleviate this problem, the output of our
  % ensemble net should be nearly binary, which requires the
  % activation
  % function of our ensemble net should has a shape similiar to step
  % function while still differentiable.
%  
%
% In this section, we describe our proposed model in detail.
% {\bf Design Justifications}
\section{Remedy: Two-stage-multi-sized FCNs and a new activation function}
As we mentioned in the previous section, when a single DSC-based FCN
is utilized to segment lesions that have a vast variability in size,
% problems may emerge, and
different types of segmentation biases may be generated.
A combined remedy would be: 1) processing the images with sub-volumes of
different sizes; and 2) treating
% As we mentioned above, the design goal of our model is to treat
large lesions and small lesions separately.
% , and replying on different % sized FCNs to best capture them.
While FCNs can handle images with arbitrary sizes, predetermined input
size does affect the design of the network, especially the number of
the layers.
FCNs with small-sized input (we term them small-FCNs) and large-FCNs
% FCNs with large-sized inputs
are more accurate for their respective sized lesions. With this in
mind, the design of the fusion step should grant small-FCNs with more
power to determine small lesions, while let large-FCNs be more
authoritative % have
% more
% to say
about large lesions.  This thought leads to the architecture of our
two-stage-multi-sized FCNs model. In the first stage, we set up three
FCN networks, for different sized neighborhoods, to best capture both
large and small lesions.  In stage two, the preliminary 3D lesions
masks, which we term {\it opinions}, are first separated into small
and large lesion groups, and then combined through nonlinear weighting
schemes carried out by ensemble-nets.
% that is learned automatically.

  % Semantic-Wise CNN Architecture This type of architecture makes
  % predictions for each pixel of the whole input image like semantic
  % segmentation [30, 31]. Similar to autoencoders, they include
  % encoder part that extracts features and decoder part that
  % upsamples or deconvolves the higher level features from the
  % encoder part and combines lower level features from the encod- er
  % part to classify pixels. The input image is mapped to the
  % segmentation labels in a way that minimizes a loss function.

  % As shown in Fig.~\ref{fig:overall_model}, our model consists of two
  % major stages. First, three 3D U-Net-like networks, are trained to
  % generate lesion segmetnations through different sized FCNs.  In the
  % stage 2, the three segmentation candidates are fused together to
  % generate the final voxel labels with enhanced accuracy and
 % consistency.
% structural details and intensity contrasts.

\begin{figure}[t]
\begin{center}
  \includegraphics[width=4in]{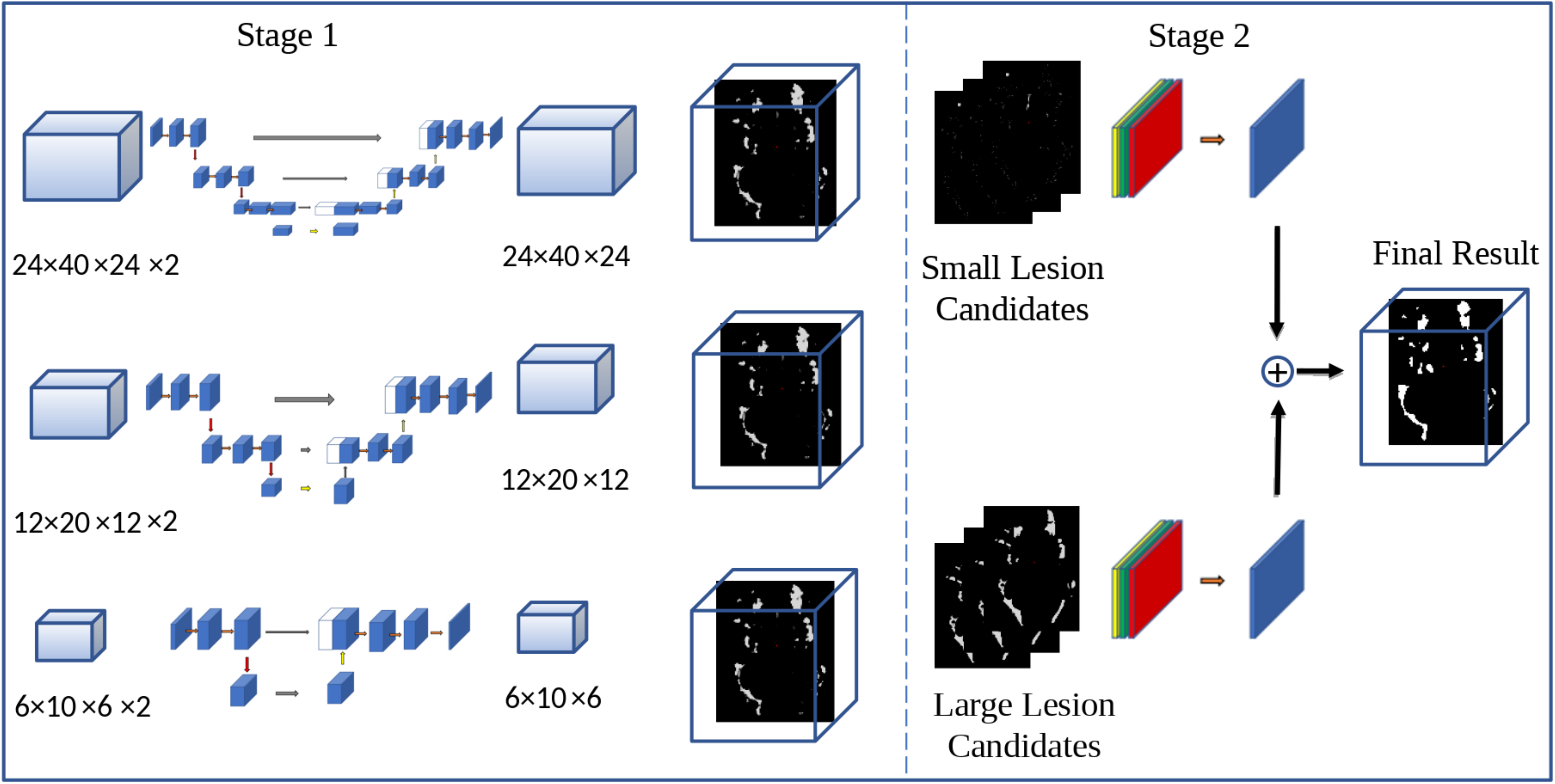}
\end{center}
\caption{The overall structure of our model for lesion segmentation.}
\label{fig:overall_model}
\end{figure}

\subsubsection{Stage 1: three different sized FCNs}
In stage 1, we extract MR patches of three different sizes, and feed
them into the corresponding FCNs. As each voxel may be covered by
multiple patches, the prediction patches of each FCN are then sent
back to the original image space to generate an overall probability
map through averaging.

In order to fit our application and data, we made modifications to the
3D U-net \cite{3d_unet} as follows. We keep the original design that
two convolution layers are followed by one pooling/upsampling
layer.  % The max-pooling layer and devolution layer are also kept as
        % the same as in 3D U-net.
The sizes of the convolution filters are set to $3 \times 3 \times 3$
except for the bottom layer of the U shape, in which the filter sizes
are set to $1 \times 1 \times 1$. Every convolution layer is padded to
maintain the  same spatial dimension. The number of layers is
reduced from 3D U-net as our input dimensions are smaller. The
modified 3D U-net-like FCNs are shown in Fig.~\ref{fig:overall_model}
Stage 1.
% {\bf At this stage, we use the Dice loss as the objective
% function}. % , which is defined as

%\begin{equation} \label{eq:1}
%L_{Dice} = 1-\frac{2\sum_i{g_iy_i}}{\sum_ig_i+\sum_iy_i}
%\end{equation}
%where $y_i\in[0,1]$ is $i$th output of the network, $g_i\in \{0,1\}$
% is the corresponding ground truth.

\subsubsection{Stage 2: ensemble of preliminary opinions} Three {\it
  opinions} are generated in Stage 1 for each MR image.  Stage 2 is
to merge them to produce an overall lesion segmentation.
% Our strategies in this stage can be described as: firstly we divided
% the segmentation $opinion$s of the firs stage as two types, small
% lesion and large lesions. Secondly we seek the solution of
% combination for each type of lesions.
As we described in a previous section, each FCN in the first stage is
more trustworthy in processing one specific type of lesions, i.e.,
small-FCNs generate more reliable segmentations for small lesions,
while large-FCNs performs better on large lesions. To take the
advantage of this fact, we separate the estimated lesions within each
prediction map into small and large lesion groups and send them into
two different ensemble nets for fusion. In this work, an empirical
threshold of $1000$-voxel has been used to decide the category (small
or large) for any given lesion (a 3D connected
component). % belongs to small or large category.
% differentiate the two lesion groups.
% Several methods can be applied for dividing lesions as small and
% large, such as rescale all the lesions volumes between [0,1], then
% choose a certain percentage as a shreshold for separation.  In this
% paper, we simply use the volume of a lesion as measurement and
% choose 1000 pixels as a hard threshold to differentiate lesions.
\subsubsection{Ensemble net} The functionality of the two ensemble
nets can both be expressed as
\begin{equation} \label{eq:2}
 \min_{w_1,w_2, w_3} D(\mathbf{y}, f(w_1\mathbf{x_1}+w_2\mathbf{x_2}+w_3\mathbf{x_3}))
\end{equation}
where $\mathbf{x_1}$, $\mathbf{x_2}$, $\mathbf{x_3}$ are the
contributing $opinion$s from the three FCNs; $\mathbf{y}$ is the
corresponding ground truth. % $D(\mathbf{a},\mathbf{b})$
$D(\cdot,\cdot)$ measures the distance between the combined opinion
and the ground-truth. With different approaches for the combination,
$f(\mathbf{x})$ can be formulated as either a linear or non-linear
function that maps $\mathbf{x}$ to the probability space.  As we
explained before, the combining weights $w_1$, $w_2$ and $w_3$ should
be individualized for small and large lesion groups,
respectively. Learning such weights can also be conducted through
neural networks.
%where ${x_i}_1$,${x_i}_2$,${x_i}_3$ are three $opinion$s at voxel $i$;
%${y_i}$ is corresponding ground truth. Depending on the design of the
% combination, $f$ can be either a linear or non-linear function. As we
% explained before, the combining weights $w_1$, $w_2$ and $w_3$ should
% be individualized for small and large lesion groups, respectively. Again,
% learning of these weights can be carried out through neural networks.
% We observed that this problem can be solved under deep learning
% framework.
In both small and large lesion groups, if we concatenate the three
{\it opinions} as three channels, and treat $w_1$, $w_2$ and $w_3$ as
a filter of size $(1,1,1,3)$, we can build an ensemble net of one
convolution layer to learn the optimal weights
%. and apply backpropagation to find a solution for
in Eqn.~\ref{eq:2}. The $f(\mathbf{x})$ function in Eqn.~\ref{eq:2}
works as the activation function for this ensemble net.

We use Dice loss as the objective function of our ensemble-nets, to
produce probability maps that match the ground-truth binary
segmentations.
% of the networks are binary segmentations.
As we explained in section \ref{section2}, a theoretical gap exists
between the DSC and Dice loss.
% , and activiation function can play an important role for a
% well-constructed approximation. %, at least from
% the numerical
% perspective.
% For Dice loss,
An observation is that activation functions that are closer to the
Step function would be able to produce better approximations for the
DSC, at least from the numerical
point-of-view.  %In other words, steep activation functions are generally
% preferreed in this regard.  On the other hand, such steep function,
% if to be used as an activation function, should have a resonable
% capture range to facilitate convergence.
In this regard, an ``ideal'' activation function should be
differentiable and relatively steep around $0.5$. Meanwhile, a
reasonable capture or support range should be allowed to avoid
vanishing gradients.
  %                                                                                                
% from the convergence considertation, functions with very narrow
% capture range should also be avoided.  {\bf With these two
% considerataions, I propose a new activitation function for our
% ensemble net.}

%As Dice coefficient is ususally used as a measurement to evaluate the
%final results, we also use Dice loss as equation \ref{eq:1} as loss
%function for our ensemble net.  Dice coefficient is based on binary
%values, and it is not differentiable.  To overcome this problem, in
%the Dice loss function the prediction results are probabiliity, which
%are continous values between 0 and 1.  Due to the relaxation, the loss
%function cannot refelect the change of Dice coefficient well. For
%example, consider about two totally overlapped circles, one as the
%ground truth and the intensity value is 1 while the other one as
%prediction.  If in the post processing, a step function takes 0.5 is
%used to map the probability as binary values, then as long as the
%intensity of the prediction circle is larger than 0.5, the Dice
%coefficient should be 1.  However, the Dice loss is 0.8235 when the
%prediction intensity is 0.7, while the Dice loss
%is 88.89 when the prediction intensity is 0.8.\\
%To narrow the gap between Dice coefficient and Dice loss, the output
%of our ensemble net should be the closer to binary the better, which
%requires the activation function of our ensemble net should satisfy
%1. the more similiar to step function the better.  2. Differentiable.

In FCN solutions for binary segmentation, the Sigmoid function is a
commonly used activation function. In our ensemble nets, however, the
inputs to the function $f(\mathbf{x})$ have a particular range of
$[0, 1]$, and a steeper function, if tailored to the inputs, would be
preferred over the Sigmoid in producing more accurate DSC
approximations. With these two considerations, we propose a new
activation function for our ensemble nets:
\begin{equation}
H(x)=\begin{cases}
    1 \quad \phantom{\infty}\text{if}\,\, x > 1  \\
    0 \quad \phantom{\infty}\text{if}\,\, x < 0 \\
    x+\frac{1}{2\pi}\sin(2\pi x-\pi) \quad \phantom{\infty}\text{if}\,\, 0\leq x \leq1
\end{cases}
\end{equation}

% \vspace{-0.3in}

%\begin{figure}
%\begin{center}
%  \includegraphics[width=3.5in]{images/activation.pdf}
%\end{center}
%\caption{The comparison of Sin-activation function and sigmoid}
%\label{fig:activation}
%\end{figure}

%\vspace{-0.1in}
\begin{wrapfigure}{r}{2.3in}
  \includegraphics[width=2.3in]{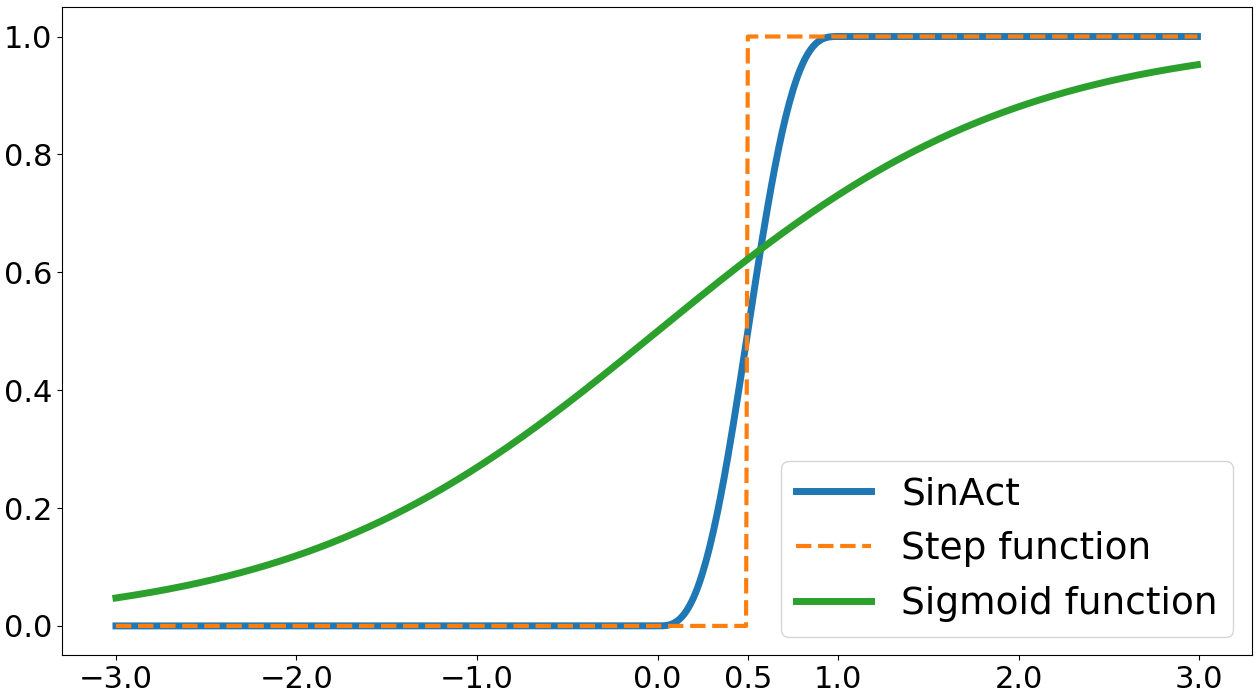}
  \vspace{-0.1in}
  \caption{Comparison of Step, SinAct and Sigmoid functions.}
\label{fig:activation}
\end{wrapfigure}

We term $H(x)$ SinAct function. Fig.~\ref{fig:activation} shows a
comparison of the Step, SinAct and Sigmoid functions. Compared with
Sigmoid, the SinAct is steeper between $0$ and $1$. As we usually use
$0.5$ as the threshold to convert probability maps into binary
results, being steep around $0.5$ is a desired property. Another merit
of SinAct function is that it evaluates to $0$ when the input is equal
or smaller than $0$, and produces $1$ when the input is equal to $1$
and beyond. By contrast, the Sigmoid can approach infinitely close to
$0$ and $1$, but would never reach the exact values.
% or larger than $1$.
% The Sigmoid does not have this property.
% For sigmoid, the function is $f(x) = 1/(1+e^{Dx})$. By changing the
% value of D we also can make the curve of sigmoid steeper. However,
% as the output of sigmoid function only can close to 0 and 1, it is
% still not good enough.

%We should note that the strategy use deep neural networks to solve the
%optimization in \ref{eq:2} can be extended to more general cases.  If
%adjacent context information of each pixel are needed for combination,
%we can use the structure of this single-layer network, and simply
%adjust the dimenions of the filers.  For example, if we want to use
%surrounding $5\times5\times5$ pixels, with $3$ probability maps for
%merge, the filter is defined as a shape of $(5,5,5,3)$.  With the nice
%numerical properties we laied out, the proposed Sin-activation
%function has the a great poential to be used as a general-purpose
%activation function in many deep learning solutions.

%\input exp_short

\section{Experiments} % al results}
%\subsection{Data and Implementation details}
{\bf Data} The MR data used in this paper were obtained from the White
Matter Hyperintensities (WMH) Segmentation Challenge at MICCAI 2017
(\url{ http://wmh.isi.uu.nl/}). The training set of this challenge
contains images of 60 subjects acquired at three sites \cite{li2018fully}.
% 20 of them were scanned at UMC Utrecht with 3T Philips Achieva, 20 at
% NUHS Singapore with 3T Siemens TrioTim and 20 at VU Amsterdam with 3T
% GE Signa HDxt.
%
For each subject, a pair of aligned T1-weighted and Fluid-attenuated
inversion recovery (FLAIR) images are available. Several
pre-processing steps are conducted on all image pairs. First, skulls
are removed from both T1 and FLAIR images using FSL/BET tool. Second,
the intensities of all images are individually normalized to the range
of $[0,1]$.  3D patches (sub-volumes) are extracted from the
intensity-normalized brains using sliding windows. The patch sizes are
set to $6\times10\times6$, $12\times20\times12$ and
$24\times40\times24$ respectively, and the strides are half of the
corresponding patch sizes. T1 and FLAIR patches from the same areas
are concatenated as two channels and fed into the FCNs in Stage 1.

{\bf Implementation Details} With the sliding windows of the
aforementioned patch sizes, we found the extracted patches containing
no lesion at all (``empty patches'') greatly outnumbered those with
lesions inside. This would create a serious data imbalance in the
training if we would use all the patches.
% To avoid the imbalance issue and make the training samples relative balanced,
To tackle this issue,
% between lesion and empty patches, %in the traning step of the Stage 1
we kept all the lesion patches, but only randomly picked an equal
number of empty patches.
% picked all the lesion patches in the training step, but only
% that contain lesions, while
% randomly selected an equal number of empty patches to make the
% training samples relatively balanced.

The three FCNs in Stage 1 are developed under Keras. ReLu is used as
the activation function for all layers except the last one, where
softmax is utilized to produce the final prediction. Dice loss is used
as the network objective function and ADAM is adopted as the
optimization method.
% For the FCNs with input patch sizes of
% $6\times10\times6$, $12\times20\times12$ and $24\times40\times24$,
% there are 831,714, 2,650,978 and 11,204,706 trainable parameters
% respectively.
The learning rate is set to $5\mathrm{e}{-6}$, %$5e-6$,
and it reduces $50\%$ every
10 epochs. In total, each FCN is trained for 30 epochs.
In Stage 2, lesions are separated into two groups, small and large
lesions, and FCN ensembles are carried out within the individual
groups. Volume size of $1000$ voxels has been used as a hard threshold
to determine the category for each 3D connected component.
% to divide the lesions into small candiates and large candidates.
For the training subjects, the separation is based on their
corresponding ground-truth, which means the component on the
probability maps is assigned as a large lesion candidate if the
corresponding groud truth component is larger than $1000$ voxels. In
the testing stage, a threshold of $0.5$ is applied on the prediction
probability map to produce a binary component first, then the map is
assigned as either small or large based on the same threshold. To
train our ensemble net, all three parameters in Eqn.~\ref{eq:2} are
initialized to $1/3$. Each ensemble net is trained for $10$ epochs.

Our experiment is carried out based on the training set of the WMH
Segmentation Challenge through Monte Carlo cross-validation. $54$ of
the $60$ subjects are randomly selected for training and the rest $6$
subjects are used for testing. The whole experiments are repeated $5$
times. % with Monte Carlo cross-validation method.
The results are presented in next section.

\subsection{Experimental results}

Evaluations of the individual components and overall network are
carried out based on five performance metrics: {\it
  DSC}, %%coefficient},
{\it Hausdorff distance} (HD, 95th percentile), {\it Average volume
  difference} (AVD, in percentage), {\it Sensitivity for individual
  lesions} (Detection, in percentage) and {\it F-1 score for
  individual lesions}.  The evaluation code was provided by the
Challenge organizers, available under the Challenge website.

The final segmentation results are shown in
Table~\ref{T:all_lesions}. FCNs with fixed input sizes are the
baseline models, and we take ensemble-nets through {\it majority vote}
and Sigmoid as the competing ensemble solutions. % the baseline model.
Our SinAct-based ensemble net outperforms all other models
% i.e., FCNs with fixed input sizes and ensemble-nets through voting
% and Sigmoid,
in every evaluation metric except detection. In detection, FCNs with
the smallest patch size of $6\times10\times6$ achieve the highest
score, but they perform the worst in all other metrics, in part
because of high false positive (type I) errors. Compared with {\it
  vote} and Sigmoid, our SinAct-based ensemble-net solution achieves
higher accuracy (measured by DSC) and consistency (measured by HD and
AVD) at the same time.

\vspace{-0.1in}
% \begin{table}[!ht]
\begin{table}[t]
  \caption{Segmentation results on all lesions.}
\centering
\scalebox{0.9}{
\begin{tabular}{c|c|ccccc}
  \hline
  \hline
%\multicolumn{1}{c}{}&
\multicolumn{1}{c|}{}&\multicolumn{1}{c|}{}&
\multicolumn{5}{c}{\textbf{Results}}\\ 
\cline{3-7}
% \multicolumn{1}{c}{\textbf{Classifier}}&
\multicolumn{1}{c|}{\textbf{FCN}} &
\multicolumn{1}{c|}{\textbf{Ensemble}}&
\multicolumn{1}{c}{\textbf{Dice}} &
\multicolumn{1}{c}{\textbf{HD}} &
\multicolumn{1}{c}{\textbf{AVD}}&
\multicolumn{1}{c}{\textbf{Detection}}&
\multicolumn{1}{c}{\textbf{F1}}\\ 
\hline

  \multicolumn{1}{c|}{\text{ $6\times10\times6$}}& & 77.74&17.82&21.72&\bf{81.83}&57.41 \\
  \cline{1-7}
  \multicolumn{1}{c|}{\text{ $12\times20\times12$}}&  & 79.78&4.65&19.28&72.43&69.91 \\
  \cline{1-7}
  \multicolumn{1}{c|}{\text{ $24\times40\times24$}}&  & 80.39&3.95&18.43&73.55&71.03 \\
  \cline{1-7}
  \multicolumn{1}{c|}{\text{3 FCNs}}& \multicolumn{1}{c|}{\text{Vote}} & 81.03&3.95&18.43&73.55&71.03 \\
  \cline{1-7}
  \multicolumn{1}{c|}{\text{3 FCNs}}& \multicolumn{1}{c|}{\text{Sigmoid}} & 80.96&4.61&19.90&80.24&68.13 \\
   \cline{1-7}
  \multicolumn{1}{c|}{\text{3 FCNs}}& \multicolumn{1}{c|}{\text{SinAct}} & \bf{81.26}&\bf{2.58}&\bf{17.54}&73.70&\bf{71.65} \\
  \hline
\end{tabular}
}
\label{T:all_lesions}
\end{table}

%FCNs with the smallest patch achieve the best score in detection, for
%both small and large lesions. However, due to the false positive (type
%I) errors, it performs the wrost in Dice, HD and AVD. Compared with
%voting, our ensemble net with SinAct can increase {\bf Dice} and
%improve consistancy, at the same time keep a high detection score.

\begin{table}[!ht]
% \begin{table}[!hbp]
%\vspace{-2mm}
  \caption{Model performance on small and large lesions separately. }
\centering
%\vspace{-1mm}
\scalebox{0.9}{
\begin{tabular}{c|c|c|ccccc}
  \hline
  \hline
%\multicolumn{1}{c}{}&
\multicolumn{1}{c|}{}&\multicolumn{1}{c|}{}&\multicolumn{1}{c|}{}&
\multicolumn{5}{c}{\textbf{Results}}\\ 
\cline{4-8}
% \multicolumn{1}{c}{\textbf{Classifier}}&
\multicolumn{1}{c|}{\textbf{FCN}} &\multicolumn{1}{c|}{\textbf{Lesion Size}}&
\multicolumn{1}{c|}{\textbf{Ensemble}}&
\multicolumn{1}{c}{\textbf{Dice}} &
\multicolumn{1}{c}{\textbf{HD}} &
\multicolumn{1}{c}{\textbf{AVD}}&
\multicolumn{1}{c}{\textbf{Detection}}&
\multicolumn{1}{c}{\textbf{F1}}\\ 
\hline

  \multicolumn{1}{c|}{\text{ $6\times10\times6$}}& \multicolumn{1}{c|}{\text{Small}} & & 59.30&18.06&47.87&\bf{80.79}&55.85 \\
  \cline{1-8}
  \multicolumn{1}{c|}{\text{ $12\times20\times12$}}& \multicolumn{1}{c|}{\text{Small}} & & 60.90&10.24&29.55&70.36&67.39 \\
  \cline{1-8}
  \multicolumn{1}{c|}{\text{ $24\times40\times24$}}& \multicolumn{1}{c|}{\text{Small}} & & 61.20&9.35&29.67&64.22&66.04 \\
  \cline{1-8}
  \multicolumn{1}{c|}{\text{3 FCNs}}& \multicolumn{1}{c|}{\text{Small}} &\multicolumn{1}{c|}{\text{Vote}} & 63.64&9.63&28.69&71.93&69.52 \\
  \cline{1-8}
  \multicolumn{1}{c|}{\text{3 FCNs}}& \multicolumn{1}{c|}{\text{Small}} &\multicolumn{1}{c|}{\text{Sigmoid}} & 65.07&11.51&34.42&78.98&64.74 \\
  \cline{1-8}
  \multicolumn{1}{c|}{\text{3 FCNs}}& \multicolumn{1}{c|}{\text{Small}} &\multicolumn{1}{c|}{\text{SinAct}} & \bf{65.80}&\bf{8.00}&\bf{26.52}&72.47&\bf{70.34} \\
  \hline
  \hline
  \multicolumn{1}{c|}{\text{ $6\times10\times6$}}& \multicolumn{1}{c|}{\text{Large}} & & 85.59&7.08&15.33&\bf{98.89}&\bf{90.02} \\
  \cline{1-8}
  \multicolumn{1}{c|}{\text{ $12\times20\times12$}}& \multicolumn{1}{c|}{\text{Large}} & & 84.92&6.40&14.10&77.87&66.54 \\
  \cline{1-8}
  \multicolumn{1}{c|}{\text{ $24\times40\times24$}}& \multicolumn{1}{c|}{\text{Large}} & & 85.13&6.81&12.99&69.39&64.22 \\
  \cline{1-8}
  \multicolumn{1}{c|}{\text{3 FCNs}}& \multicolumn{1}{c|}{\text{Large}} &\multicolumn{1}{c|}{\text{Vote}} & 85.32&8.23&14.05&76.31&68.19 \\
  \cline{1-8}
  \multicolumn{1}{c|}{\text{3 FCNs}}& \multicolumn{1}{c|}{\text{Large}} &\multicolumn{1}{c|}{\text{Sigmoid}} & \bf{86.87}&\bf{5.81}&12.58&88.59&84.66 \\
  \cline{1-8}
  \multicolumn{1}{c|}{\text{3 FCNs}}& \multicolumn{1}{c|}{\text{Large}} &\multicolumn{1}{c|}{\text{SinAct}} & 86.69 & 5.87&\bf{12.33}&86.84&76.84\\
  \hline
\end{tabular}
}
%\vspace{2mm}
% Boldface denotes the best performance for each measure.}
%\vspace{-10mm}
\label{T:small_large}
\end{table} 

Table~\ref{T:small_large} provides evaluations of the models on small
and large lesions separately.
% Similar to the all-lesions experiment, the FCNs with smallest patch
% size obtain the best results in detection, but does not perform well
% in DSC.  that divided lesions as small and large. The same as the
% performance on the whole lesions, the FCNs with patch size
% $6\times10\times6$ achieved the best detection results on small
% lesions, but doesn't perform well on Dice score.
For both small and large lesion groups, almost all ensemble nets
outperform the fixed-sized FCNs in DSC, which to certain extent
validates our design of combining multi-sized FCNs. As our SinAct
function provides more accurate DSC approximations at voxels with
uncertain labels, it improves the membership decision most
significantly at the boundary voxels.
% the better DSC approximations brought by our SinAct function are
% more significant at voxels with uncertain membership, it improves
% the label decisions mostly along the boundary voxels.  our SinAct
% function provides more accurate DSC approximations than Sigmoid, it
% brings improves the
Among the three ensemble-nets, our SinAct net performs the best, in
all metrics, for small lesions. For large lesions, SinAct is still
better than {\it vote}, but achieves comparable performance with the
Sigmoid. 
% achieve comparable performance, both better than that of {\it vote}.
This disparity can be explained by the fact that boundary voxels
account for high percentage for small lesions, but not so for large
lesions.
% An possible explanation for this disparity is that our SinAct
% improves label decisions mostly along the boundary voxels, which
% account for high percentage for small lesions, but not so for large
% lesions. All in all, our two-stage-multi-sized
All in all, FCNs and the new activation function are shown to be
effective in improving MS lesions segmentations from MRIs.

% For ensemble net, although the improvement on Dice Coefficient is
% $0.2\%$ for the whole lesions when compare with baseline, it can
% increase Dice Coefficient on small lesions and large lesions by
% $2\%$ and $1.3\%$ respectively, while still can improve consistency
% of the results and keep a high detection score.  How many sentences
% can I add?  How many sentences can I add?  How many sentences can I
% add?

%\begin{itemize}
%\item 6 -> 12 -> 24: detections are both decreasing. Missing small lesions. 
%\item HD smaller: consistency.
%\end{itemize}

%\begin{figure}[!htb]
%\begin{center}
%  \includegraphics[width=3.2in]{images/present_results.png}
%\end{center}
%\caption{Upper row is a subject with lesions consists of large and
% small components; lower row is a subject with lesions that only
% consists of small components. For each row, the four images are
% FLAIR, T1, Ground Truth and our segmentation results
% correspondingly.}
%\label{fig:results}
%\end{figure}

%\input con_short
\section{Conclusions}

In this paper, we propose a two-stage-multi-sized FCNs strategy to
enhance the segmentation of MS white matter lesions in MR images. The
design is based on the rational that different sized lesions are best
captured with appropriate sized FCNs. Ensemble-nets are constructed to
combine the results from the FCNs, where SinAct, a new activation
function, is adopted to improve the segmentation accuracy measured by
DSC. Experiments show the effectiveness of both design
approaches. Exploring more activation functions is the directions of
our future efforts. We are also interested in applying the proposed
strategy to other neuroimage analysis problems.

\bibliographystyle{splncs03}
\bibliography{mlmi_2018}

\end{document}